%% file: submission.tex
\pdfoutput=1

\documentclass[11pt]{article}

\usepackage[preprint]{acl}

\usepackage{times}
\usepackage{latexsym}
\usepackage{booktabs}
\usepackage{longtable}
\usepackage{geometry}
\usepackage[T1]{fontenc}

\usepackage[utf8]{inputenc}

\usepackage{microtype}

\usepackage{inconsolata}

\usepackage{graphicx}

\usepackage{bbding}
\usepackage{xcolor}
\newcommand{\cm}{\textcolor{green}{\Checkmark}}
\newcommand{\buhbuh}{\textcolor{red}{\XSolidBrush}}
\newcommand{\bad}{\textcolor{red}{$\uparrow$}}

\newcommand{\good}{\textcolor{green}{$\downarrow$}}
\newcommand{\vgood}{\textcolor{green}{$\downarrow\downarrow$}}

\newcommand{\cbtwo}{\texttt{clembench}$_{2}$}

\usepackage{enumitem}
\setlist[description]{font=\normalfont\itshape\space}

\usepackage{colortbl}
\usepackage{makecell}
\usepackage{multirow}
\usepackage{supertabular}

\title{A Third Paradigm for LLM Evaluation:\\ Dialogue Game-Based Evaluation using \texttt{clembench}}

\author{David Schlangen\textsuperscript{1,2} $\,$ Sherzod Hakimov\textsuperscript{1} $\,$ Chalamalasetti Kranti\textsuperscript{1}\\ \textbf{Jonathan Jordan\textsuperscript{1} $\,$ Philipp Sadler\textsuperscript{1}}\\
        \textsuperscript{1} Computational Linguistics, University of Potsdam, Germany\\
        \textsuperscript{2} DFKI (German Research Center for AI), Berlin}

\input{latex/cbd-preamble}

\begin{document}
\maketitle
\begin{abstract}
There are currently two main paradigms for evaluating large language models (LLMs), \textit{reference-based evaluation} and \textit{preference-based evaluation}. 
The first, carried over from the evaluation of machine learning models in general, relies on pre-defined task instances, for which reference task executions are available.
The second, best exemplified by the LM-arena, relies on (often self-selected) users bringing their own intents to a site that routes these to several models in parallel, among whose responses the user then selects their most preferred one.
The former paradigm hence excels at control over what is tested, while the latter comes with higher ecological validity, testing actual use cases interactively. 
Recently, a third complementary paradigm has emerged that combines some of the strengths of these approaches, offering control over multi-turn, reference-free, repeatable interactions, while stressing goal-directedness: \textit{dialogue game based evaluation}. While the utility of this approach has been shown by several projects, its adoption has been held back by the lack of a mature, easily re-usable implementation. In this paper, we present \texttt{clembench}, which has been in continuous development since 2023 and has in its latest release been optimized for ease of general use.
We describe how it can be used to benchmark one's own models (using a provided set of benchmark game instances in English), as well as how easily the benchmark itself can be extended with new, tailor-made targeted tests.\footnote{All code required to run the benchmark, as well as extensive documentation, is available at \url{https://github.com/clembench/clembench}.}
\end{abstract}

\section{Introduction}

\begin{table}[]
    \centering
    {\small
    \begin{tabular}{r|c|c|c}
      feature / \textit{paradigm}    &     \textit{rb }      &     \textit{pb}       &    \textit{dgb} \\ \hline \hline
    control over task   & \cm   &   \buhbuh   & \cm  \\ \hline
    replicability & \cm   &   \buhbuh   & \cm  \\ \hline
    turns              &   1  & 1+  & 1+   \\ \hline    
    parity text / image &   \buhbuh   &  \buhbuh   & \cm  \\ \hline
    cost of benchmarking &  fixed   &  open   & fixed  \\ \hline
    leakage danger &  \bad & \vgood  &  \good  \\ \hline
    extension cost (instances) &  \bad &   open &  \vgood  \\ \hline
    extension cost (tasks) &  \bad &  no control &  \good  \\ \hline
    saturation danger &   \bad & \good & \good \\ \hline
    \end{tabular}
    }
    \caption{A feature matrix comparing \textit{reference-based (rb)}, \textit{preference-based (pb)} and \textit{dialogue game-based (dgb)} evaluation paradigms across dimensions}
    \label{tab:matrix}
\end{table}

When a new large language model is released, it is customary now to provide two types of evaluation results (see, e.g., \cite{team_gemma_2025}): First, results are given from user rankings, as collected on the LM Arena (formerly Chatbot Arena, \citet{chiang_chatbot_2024}). This is meant to represent general user perception of model quality. Second, results on ``standard benchmarks'' are provided, for example such as are collected by the LLM-harness \cite{eval-harness}. This, in turn, is meant to measure specific capabilities in more depth and with more control than is brought out by the user preferences. While these two types of evaluation complement each other (controlled and static, less controlled and interactive), it has recently been argued that a third combination (controlled and interactive) is needed to acquire a well-rounded impression of the capabilities of a model \cite{momentè2025triangulatingllmprogressbenchmarks}. 

As discussed below (Section~\ref{sec:relwo}), there have been several attempts at realising controlled, interactive evaluation through making LLMs self-play conversational games. Arguably, widespread adoption of this paradigm has been held back by the fact that many of these attempts appear to have been built for a single evaluation campaign only, with less attention being paid to re-usability of code. The package highlighted here, \clembench\ \cite{chalamalasetti-etal-2023-clembench}, was one of the first instantiations of this idea, and has driven a leaderboard that has been continuously updated since early 2023. Thanks to a grant by the University of Potsdam, we can now offer \cbtwo, which has been optimized for ease of use in various use cases, including local evaluation of LLMs or LLM fine-tunes. In this paper, we first briefly describe the ``dialogue game-based evaluation'' paradigm in general, explaining how it differs from the other paradigms (as illustrated in the feature matrix in Table~\ref{tab:matrix}). We then focus specifically on \cbtwo, with the use case of evaluating new model in mind. For this, we show how the existing benchmark can be run, to yield results that can directly be compared to the long history of measurements recorded on the official leaderboard.\footnote{
    \url{https://clembench.github.io/leaderboard.html}
}
We also show how easily the benchmark can be extended with games highlighting special abilities.\footnote{%
    The scientific basis behind the approach is discussed, evaluated, and validated at length in \cite{chalamalasetti-etal-2023-clembench,hakimov-etal-2025-using}. The use of the framework as an environment for reinforcement learning of multi-turn problems has been explored by \citet{horst2025playpenenvironmentexploringlearning}.
    The present paper aims at making the community aware of it as a resource for easy evaluation of new models or model checkpoints during development.
}

\section{Evaluation Through Self-Play of Dialogue Games}

\subsection{The General Idea}
\label{sec:genid}

\begin{figure}
{ \small
\begin{description}[noitemsep]
   \item[A (quietly reads game card):] \texttt{Target word: ugly. Taboo words: displeasing, despicable, unattractive}.
   \item[(1a) A (to B):] Something that is not pleasing to the eye, often making a bad impression.
   \item[(1b) B:] Eyesore
    \item[(2a) A:] More about the quality of a person or thing, rather than a nuisance or distraction.
    \item[(2b) B:] Ugly.
    \item[(3a) A:] Yay! Well done us. 
    \end{description}
        \vspace*{-.5cm}
}
    \caption{A constructed example of a successful episode of the two-player Dialogue Game \texttt{Taboo}, as played by humans}
    \label{fig:taboo-hh}
\end{figure}

\begin{figure}
{ \small
\begin{description}[noitemsep]
   \item[Game Master (GM) to A:] \textcolor{gray}{We are playing a collaborative word guessing game. Your task is to describe a concept, without using its name, and without using some other related terms. The target concept is: ugly; the related words are: displeasing, despicable, unattractive. Start with "CLUE: ", and be brief.}
   \item[(1a) A to GM:] CLUE: Something that is not pleasing to the eye, often making a bad impression.
   \item[(1a') GM to B:] \textcolor{gray}{We are playing a word guessing game. You need to guess a target word that another player is describing to you. You can make one guess at each trial. After each trial you will get another hint. Start with "GUESS: ", and only give a single word. The other player gave the following clue: Something that is not pleasing to the eye, often making a bad impression.}
   \item[(1b) B to GM:] GUESS: Eyesore
    \item[(1b') GM to A:] \textcolor{gray}{Your partner guessed: ``eyesore''. Please provide another clue, starting with "CLUE:".}
    \end{description}
    \vspace*{-.5cm}
}
    \caption{Llama-3.1-405B in prompted self-play of the same instance of \texttt{Taboo} as in Figure\ref{fig:taboo-hh} (excerpt).}
    \vspace*{-.5cm}
    \label{fig:taboo-sp}
\end{figure}

Figure~\ref{fig:taboo-hh} shows an example of the kind of game that has been used in the "dialogue game-based evaluation paradigm". In this (constructed) example, two humans are playing the game "taboo", in which a \textit{describer} has to provide clues about a concept to a \textit{guesser}, while following the constraint of not using the concept name and some related terms. 

\paragraph{What makes games like these useful evaluation instruments?} 
To address this question, we can see that even this simple interaction challenges certain capabilities: understanding of game rules and how they constrain the available actions; understanding of a target concept sufficiently to be able to describe it; ability to construct utterances under negative constraints (avoiding certain lexical items); decoding a clue; ability to integrate repeated clues. As has been argued in the literature discussed below, these are capabilities that static datasets such as MMLU-Pro \cite{mmlupro2024}, Big-bench Hard \cite{suzgun-etal-2023-challenging}, and IFEval \cite{ifeval} do not, or at least not as systematically and holistically, test.\footnote{
    See also \citet{Schlangen-2023-1,Schlangen-2019-1} for an extensive discussion of this question.
} 

\paragraph{How can such games be played with LLMs?} While for human player, the acquisition of the rules of a game, making the decision to play it, and the actual game play may constitute separate activities, for LLMs, these need to be folded into one. Figure~\ref{fig:taboo-sp}, which is an actual instance of self-play of an LLM,\footnote{%
    Specifically, it is LLama-3.1-405B playing;
    \url{https://github.com/clembench/clembench-runs/blob/main/v2.0/Meta-Llama-3.1-405B-Instruct-Turbo-t0.0--Meta-Llama-3.1-405B-Instruct-Turbo-t0.0/taboo/2_low_en/episode_2/transcript.html}
}
shows how \textit{prompted self-play} can be set up with such models. The crucial element here is the introduction of a \textit{Game Master} which mediates the interaction by prompting each player role into existence. In the figure, one can see how the `scaffolding' shown in a lighter font weight supports the game play (darker weight), which taken on its own is what looks more like the human/human play from Figure~\ref{fig:taboo-hh}.

\ \\
Below, we follow \citet{chalamalasetti-etal-2023-clembench} in making a distinction between the \textit{game} itself (e.g., \texttt{Taboo}), the way it is explained to a model through prompts (which we call the \textit{dialogue game instantiation}), the specific \textit{game instance} that is to be played (e.g., \texttt{Taboo} with a specific target word and set of taboo words), and the \textit{experiment} which collects a set of instances.

\subsection{Some Unique Properties}

What makes \textit{dialogue game-based evaluation} (DGBe) special? For Table~\ref{tab:matrix}, we have compiled into a feature matrix some of the properties as claimed in the work cited below. As can be seen, it is the combination of properties that makes DGBe special. Through the selection of games, instantiations, and game instances as non-changing, pre-constructed elements, a high degree of control over the task is conferred, while the fact that by self-play, only models and no human players are involved, provides a certain degree of replicability. Unlike in static datasets, many of the games allow for or even require multi-turn interactions. The separation of \textit{game} and \textit{game instance} makes it possible to explore different ways of presenting game material; the \clembench, for example, provides, for some games, coordinated text-only and text/image variants \cite{hakimov-etal-2025-using}. The fact that pre-defined experiments are being evaluated also brings DGBe closer to \textit{reference-based evaluation} (RBe), and makes the cost of running the benchmark on a model more predictable and fixed; whereas in \textit{preference-based evaluation} (PBe), this in principle is unlimited and the results get better / more stable, the longer the model is provided and compared. The last three features mentioned in Table~\ref{tab:matrix} all relate to the relative ease with which new game instances and even new games can be created (more on that below), compared to static datasets, which need to be authored or collected and curated.

\subsection{Existing Frameworks}
\label{sec:relwo}


The idea of using game self-play for evaluation has been implemented by various frameworks in recent years. An early precursor to this was TextWorld \cite{cote2019textworld}, which however operated with a single genre of game (text adventures / interactive fiction), and with the expectation of training specialist models. Only with the advent of generalist models that can be \textit{prompted} into being specialists \cite{brown_language_2020,wei_finetuned_2021} did it become possible to implement this idea at a larger scale, for single games \cite{bertolazzi-etal-2023-chatgpts} and more generally in frameworks in which various games can be implemented \cite{chalamalasetti-etal-2023-clembench,qiao2023gameeval,gong2023mindagent,wu2024smartplay,Zhou2024-sotopia,duan2024gtbench,guertler_textarena_2025,cui2025tales}.


\subsection{\clemcore\ /\ \clembench}

\begin{figure}
    \centering
    \includegraphics[width=1\linewidth]{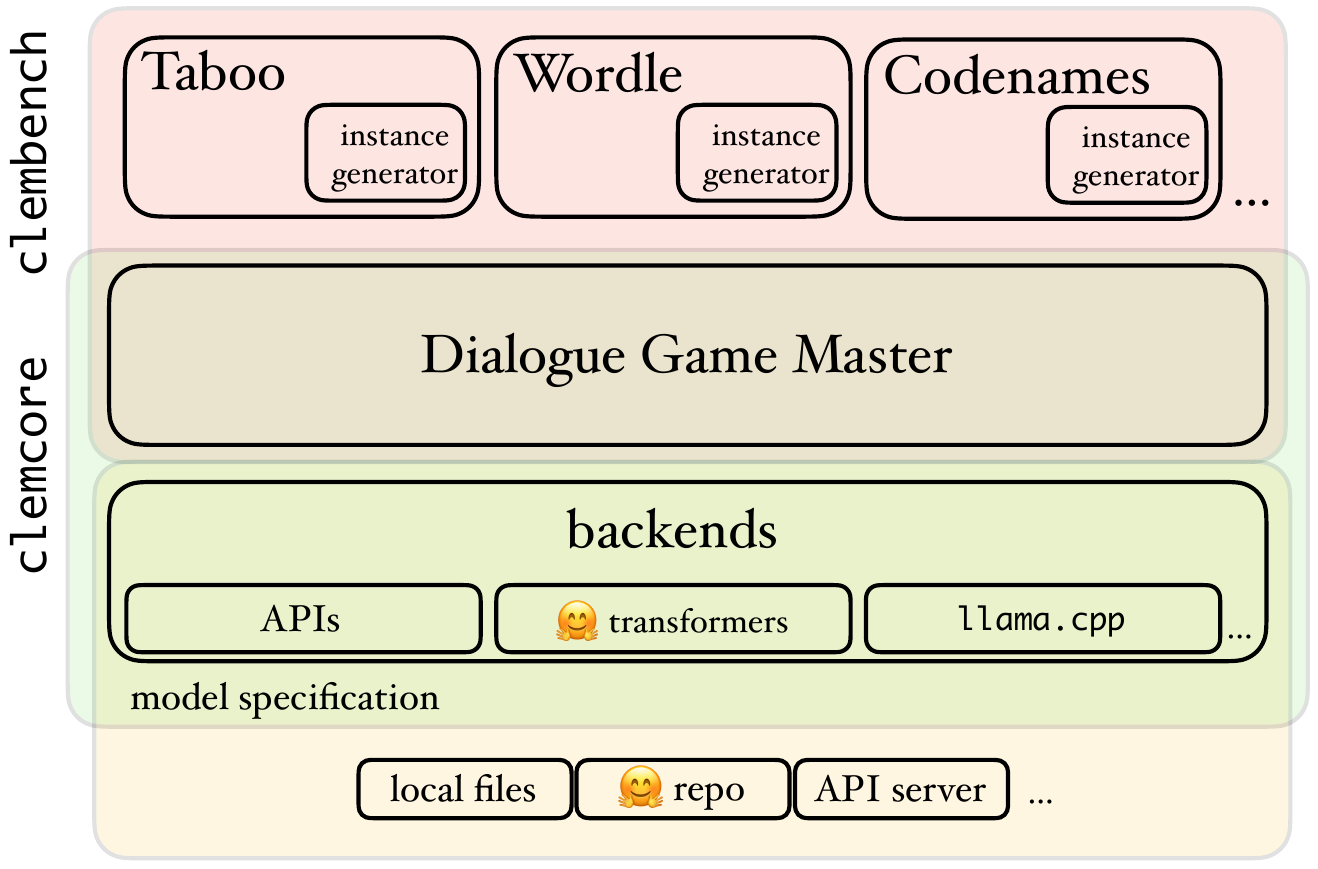}
    \caption{Schematic view of the abstraction layers provided by \clemcore\ / \clembench}
    \label{fig:abstr}
\end{figure}

The \texttt{clem} framework demonstrated in this paper has two components: A) \clemcore , which is the pip-installable backbone abstracting access to LLM into a common interface and providing the infrastructure for setting up games and running experiments on them (e.g., logging, scoring, aggregating results). B) \clembench, which is a collection of implemented games, and, for each version number of the benchmark, a different pre-compiled set of instances to run.\footnote{%
    \clemcore: \url{https://github.com/clembench/clemcore};
    \clembench: \url{https://github.com/clembench/clembench}. Both are licensed under the permissive MIT license.
}
This division of labour is illustrated in Figure~\ref{fig:abstr}.


The framework has been used to power the public `\clembench\ leader board' since 2023.\footnote{%
\url{https://clembench.github.io/leaderboard.html}
}
The long period for which this has already been run has made it possible to track interesting trends; for example, the plot in Figure~\ref{fig:trends} (Appendix) clearly shows the moment at which open-weight models started to catch up, as well as the generally fast improvements over time. (See \cite{beyer_clembench-2024_2024} for a more detailed discussion.) We also host a \textit{transcript browser} which makes it particularly easy to browse the by now quite large database of self-play transcripts of various types of proprietary and open-weight models.\footnote{%
    Transcript browser: \url{https://clembench.github.io/transcript-browser.html}. The transcripts are also all stored at \url{https://github.com/clembench/clembench-runs/} (more than 3GB by now).
}

Currently, 15 games with text-only input and 6 multimodal games (image and text inputs) are made publicly available with more being developed in the pipeline.

\section{Running \texttt{clembench}}
\label{sec:run}

Let us assume that you have trained a new LLM, or produced a new fine-tune of an existing one, and now want to evaluate it for its interactive, dialogic reasoning capabilities. How can you use \clembench\ to do that?
Detailed and up-to-date instructions can be found on the repository site,\footnote{%
    \url{https://github.com/clp-research/clembench}
}
and we only highlight some key points here.

The first decision you have to make is how you want to serve your model. Thanks to the abstraction layers described above, \cbtwo\ offers a large degree of flexibility here. Besides integrating a number of proprietary APIs (openAI, anthropic, mistral), and emerging API standards (in the backend \texttt{openAI-compatible}), the package also allows for inference via the huggingface transformers library \cite{wolf_transformers_2020}, via vLLM \cite{kwon2023efficient}, and via \texttt{llama.cpp}.\footnote{%
  \url{https://github.com/ggml-org/llama.cpp}
}
Other ways of interfacing with model inference can easily be implemented as a new backend.

In either case, the model itself is described via a structured entry in the \clembench\ \textit{model registry} which specifies all details required for inference, from the backend to use up to additional information like chat-templates and special tokens like EOS.\footnote{%
    Similarly to the newly emerging \texttt{model.yaml} approach (\url{https://modelyaml.org}), but predating it for several years. We are actively exploring possibilities for adoption.
}
This makes it possible to call the model by a provided name in the command line call to run the evaluation pipeline, with the framework automatically loading the appropriate backend and potentially also any additional files. (Figure~\ref{fig:mreg} illustrates this for a model hosted on a huggingface repository and run locally via the transformers library.)

\begin{figure}
    \centering
{\small
\begin{verbatim}
{
  "model_name": "my-model-8b",
  "huggingface_id": "my-org/My-Model-8B",
  "backend": "huggingface_local",
}
\end{verbatim}
}
    \caption{A minimal \textit{model registry} entry for a model hosted on a public repository and inferenced with the \texttt{transformers} library}
    \label{fig:mreg}
\end{figure}

The games making up the benchmark are similarly described by structured specification files (\texttt{clemgame.json}). This makes it possible to conveniently denote specific games or subsets of the whole benchmark when starting the evaluation pipeline. Completing the pipeline (inferencing, transcribing, scoring) results in detailed records of the self-play runs, consisting of transcripts (in machine-readable formats, but for convenience also as html and tex files; for the latter, see Figure~\ref{fig:tex} in the Appendix for an example), detailed statistics for all metrics specified by the individual games -- and, as the condensed form of all metrics, a single \texttt{clemscore} (between 0 and 100), which can conveniently be reported.\footnote{%
   See \url{https://clembench.github.io/leaderboard.html} to explore the various evaluation depths.
}


In the current respective latest versions (v2.0 for text-only, v1.6 for multimodal), running the whole benchmark (14 text-only games with in total 817 instances; 5 multimodal games with 560 instances), takes about 360 minutes on two NVIDIA A100 80GB GPUs for a 70B model run via the huggingface backend (text only) / 350 minutes on one A100 for a 40B model (multimodal).

\section{Extending the Benchmark} 

As mentioned above, a particularly interesting feature of game-based evaluation is the ease with which the tests can be extended or tailored to particular questions. (Of course, for \textit{comparability}, a fixed state of the benchmark must be chosen, such as that defined by the \clembench\ repository.) We briefly describe the process here, again referring the reader to the extensive documentation of the package for details.

\subsection{Creating New Instances}

For certain use cases (such as using the framework for \textit{learning}, as described by \citet{horst2025playpenenvironmentexploringlearning}), it can be necessary to create new instances of problems for the existing games. All games in \clembench\ come with an \textit{instance generator} step, which automatically creates new stimuli. How this is done differs between games, from sampling from a larger pre-compiled set (e.g., in wordle) to programmatically generating stimuli (as in imagegame). If the larger set is exhausted, creating that resource may require some manual effort (e.g., in taboo, words and related words must be derived in a semi-automatic process making use of resources such as wordnet \cite{fellbaum:wordnet}). In any case, what is never needed is the collection of reference answers -- an important point about game-based evaluation is that it is \textit{reference-free}!


\subsection{Creating New Games}

Even though the standard \clembench\ benchmark comes with a large number of games, covering a wide spectrum of capabilities \cite{beyer_clembench-2024_2024}, for certain special interests, it may be advantageous to define new games. Again, we provide extensive documentation to guide through this process,\footnote{%
    \url{https://github.com/clembench/clemgame-template}
}
of which we only cover a few highlights here. To implement the game logic, classes from \clemcore\ that implement the concept of the \textit{Game Master} (as described in Section~\ref{sec:genid} above) can be instantiated. For simple round-robin games, only scoring rules and success conditions need to be defined; more flexibility can be achieved by overwriting defaults. \clemcore\ also abstracts away all details of handling experiments and individual runs. The instance generator takes the prompt templates that describe the game to the player, together with an instance set, and creates the specifications of individual instances that are collected in an experiment. All in all, depending on the complexity, realising a game can take as little as 2-3 hours (for simpler abstract games); importantly, however, the framework also allows for the implementation of ambitous game using complex environments such as AI2-THOR \cite{kolve_ai2-thor_2022}.

\subsection{Multilingual Evaluation with \clembench: Adding New Languages}

As explained, the way a game is \textit{instantiated} so that LLMs can be made to play it is via textual descriptions on the one hand, and the game logic implemented in code on the other. The latter part is language agnostic. The \clembench\ as referred to above comes with prompts written in English, and, for games where this is relevant (e.g., wordle, taboo), with instances in English. While this is something we have only begun to explore ourselves (see initial results reported by \cite{beyer_clembench-2024_2024}), the framework is ready for defining experiments with stimuli in other languages. \citet{hakimov2025pricethoughtmultilingualanalysis} evaluated the negotiation abilities and multilingual aspect in LLMs by developing dialogue games in three languages: English, German, Italian. 
The flexibility of the machinery for running experiments (see Section~\ref{sec:run}) makes it possible to easily define language-specific as well as cross-lingual benchmark runs. We are currently setting up a mechanism for leveraging the community to collect game instantiations in a large set of languages.


\section{``Vibe-checking'' the Agentic Abilities of a Model}

A new addition in \cbtwo\ is the integration of the dialogue experiment framework \texttt{slurk} \cite{Goetze-2022-1}. This integration makes it possible to interact locally (on one's own machine) with a model as a partner in the implemented dialogue games and thus to get a feel for how it behaves in constrained agentic conditions.\footnote{%
  A common practice that has been systematically argued for e.g.\ by \cite{dunlap2025vibecheckdiscoverquantifyqualitative}.
} The \texttt{slurk} framework has also proven to be able to handle larger crowdsourcing experiments (e.g., \citet{Ilinykh-2019}), so that more thorough quantitative studies are also possible.\footnote{%
   A preview of the public interface is hosted at \url{https://clemp.ling.uni-potsdam.de}; note however that we cannot guarantee that backend models as game play partners are available at any time, so this should be seen only as a demo of the interface. The game play itself will be demonstrated at the conference.
}

\section{Evaluating the Evaluator}

\subsection{Human Game-play Performance}


To assess the validity of games and to compare with LLM performance, in \cite{beyer_clembench-2024_2024} we performed a user study on the initial set of games. We asked the participants to play 10-15 episodes per game (leaving out wordle-clue and wordle-critic, as these
are only variants of the main wordle game). All episodes were played to the end where the human participants obliged the formatting-related game rules (reaching 100\% in \textit{Played} score). The
resulting \textit{Quality} scores were as follows: wordle: \textit{72}, taboo:
\textit{80.5}; drawing: \textit{95.2}; reference: \textit{100}; leading to an average of \textit{86.93}. The best performing model \textit{o3-mini} on the \textit{clembench leaderboard (v2.0)} still lags behind the average human performance, it has an average Quality score of \textit{82.23} on the same games.

\subsection{Reproducibility of Benchmark Results}


All games used for benchmarking are publicly available as well as the their specific instance versions. We run the vanilla (without any agentic or other tools) version of benchmarked LLMs by keeping the same parameters (temperature, library/API version, etc.) across them. Thus, as long as the model providers do not make changes to their models (applies to commercial models), the benchmark results are reproducible. (To the extent that LLM calls are deterministic. However, slight differences in responses are unlikely to affect final scores, which are an average over the perfomance on a larger set of instances.)  All existing benchmark runs are made publicly available.\footnote{\url{https://github.com/clembench/clembench-runs}}

\subsubsection{The Value of Game-Based Evaluation}

\citet{momentè2025triangulatingllmprogressbenchmarks} have compared the use of games for interactive evaluation (as implemented in \clembench), the use of targetted cognitive test sets, and the use of more generalist benchmarking datasets. A central finding was that games provide a more fine-grained means to discriminate between models; the authors recommend an approach that ``triangulates'' performance through a combination of interactive and static benchmarking.


\section{Conclusions}

This paper accompanies a demonstration of the \cbtwo\ evaluation package. We hope this paper has illustrated that running game-based evaluations does not need to be complicated, and in fact, using \clembench, in most cases is a matter of running a single command (and providing a single file, describing the model that is to be evaluated). The reasons \textit{why} models should be evaluated in this way and what exactly is being measured could only be sketched here, and for a full argumentation, we refer the reader to recent works such as \cite{momentè2025triangulatingllmprogressbenchmarks} as well as the papers mentioned in Section~\ref{sec:relwo} above. 
We hope that the work reported here can contribute to further adoption of this complementary evaluation paradigm, in particular in light of the increased use of LLMs as backbones for interactive and agentic tasks.

\bibliography{custom,anthology,existing_frameworks}

\appendix
\section{Appendix}

\begin{figure*}[th!]
    \centering
    \includegraphics[width=\linewidth]{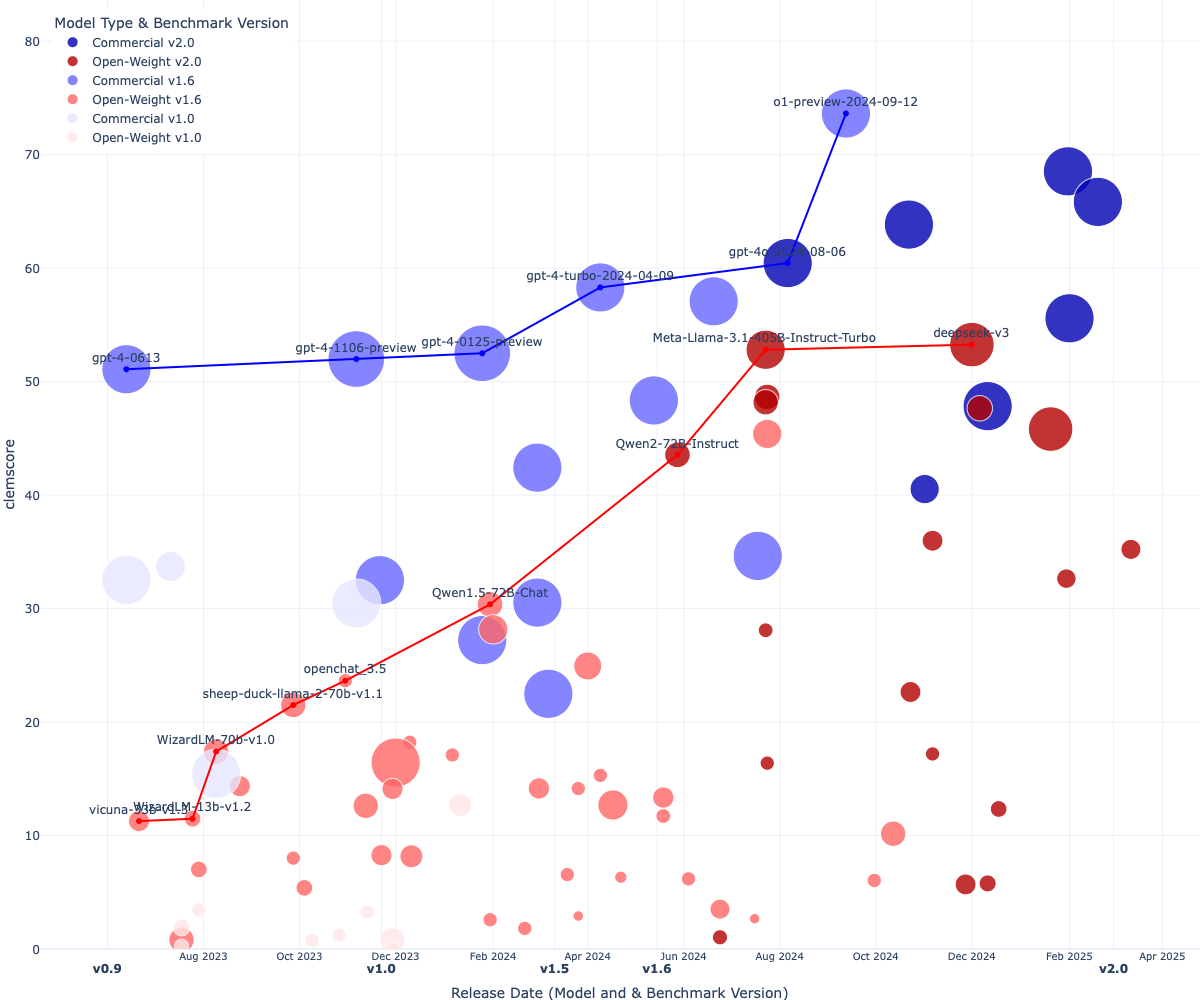}
    \caption{Development of achieved scores on clembench since 2023. Interactive version available at \url{https://clembench.github.io/leaderboard.html} ("trends").}
    \label{fig:trends}
\end{figure*}

\begin{figure*}
    \centering
    \includegraphics[width=\linewidth]{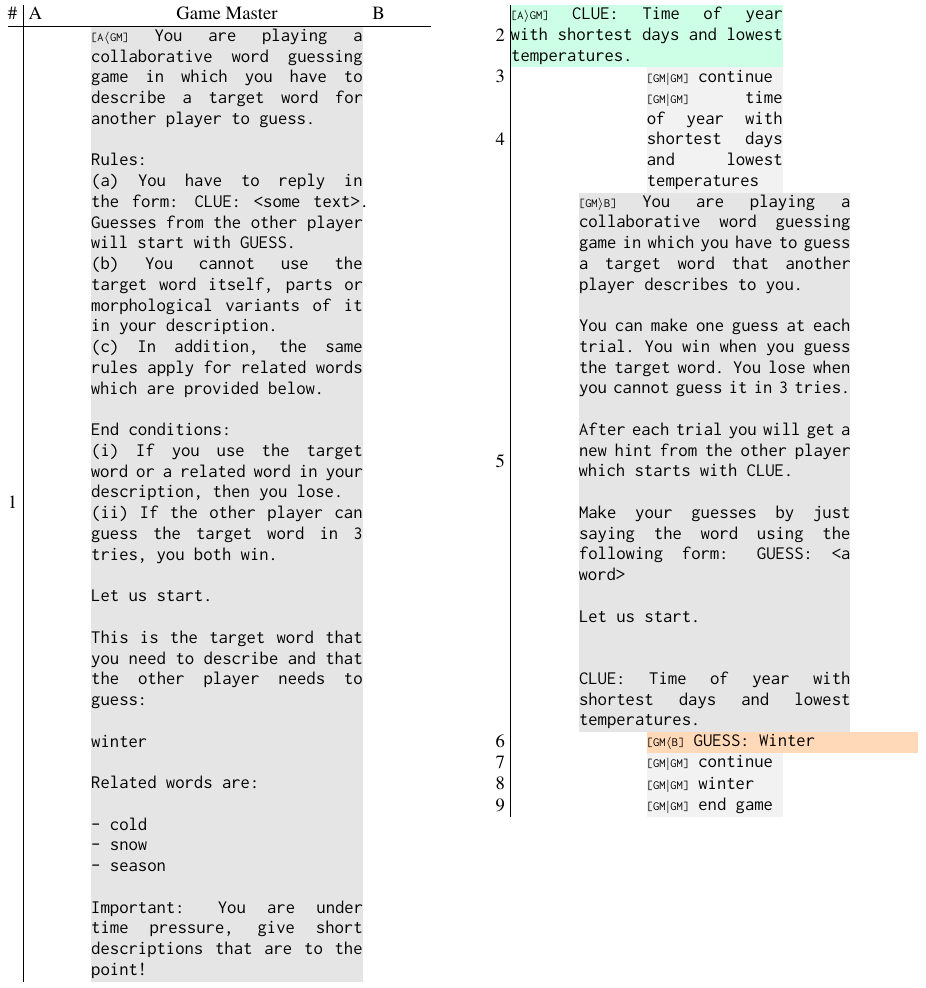}
    \caption{Example of the automatically generated transcript format. (Here: Llama-3.3-70B-instr playing Taboo, epsiode "0-high-en", 0.)}
    \label{fig:tex}
\end{figure*}

Table~\ref{tab:existing_frameworks} lists all related frameworks. The column ``Maintained'' is determined whether there have been changes on the respective Git repository in the last three months. Figure~\ref{fig:trends} shows trends observed on the \clembench\ leader board. Figure~\ref{fig:tex} shows an example transcript.

\input{existing_frameworks}



\end{document}

%% file: latex/cbd-preamble.tex
\usepackage{multicol}

\usepackage{color}

\usepackage{color}

\usepackage{booktabs}

\newcommand{\clembench}{\texttt{clembench}}
\newcommand{\clemcore}{\texttt{clemcore}}

%% file: existing_frameworks.tex
\begin{table*}[htbp]
\centering
\begin{tabular}{p{2.4cm}|p{4.8cm}|c|c|c} 
\hline
\textbf{Framework} & \textbf{Git repo} & \textbf{First commit} & \textbf{Latest commit} & \textbf{Maintained} \\
\hline
Tales \cite{cui2025tales} & \url{https://github.com/microsoft/tale-suite} & 2018-06-06 & 2025-04-23 & \cm \\
\hline
Sotopia \cite{Zhou2024-sotopia} & \url{https://github.com/sotopia-lab/sotopia} & 2023-03-31 & 2025-05-05 & \cm \\
\hline
Bytesized32 \cite{wang_bytesized32_2023} & \url{https://github.com/cognitiveailab/BYTESIZED32} & 2023-05-12 & 2024-07-08 & \buhbuh \\
\hline
Clembench \cite{chalamalasetti-etal-2023-clembench} & \url{https://github.com/clembench/clembench} & 2023-06-07 & 2025-06-30 & \cm \\
\hline

GameEval \cite{qiao_gameeval_2023} & \url{https://github.com/jordddan/GameEval} & 2023-08-13 & 2023-09-03 & \buhbuh \\
\hline

MindAgent \cite{gong_mindagent_2023} & \url{https://github.com/mindagent/mindagent} & 2023-09-19 & 2024-06-12 & \buhbuh \\
\hline
Negotiation Arena \cite{bianchi_how_2024} & \url{https://github.com/vinid/NegotiationArena} & 2023-09-30 & 2024-02-15 & \buhbuh \\
\hline
SmartPlay \cite{wu_smartplay_2023} & \url{https://github.com/microsoft/SmartPlay} & 2023-10-02 & 2024-04-11 & \buhbuh \\
\hline

OSWorld \cite{xie_osworld_2024} & \url{https://github.com/xlang-ai/OSWorld} & 2023-10-16 & 2025-07-02 & \cm \\
\hline
AgentBench \cite{liu_agentbench_2023} & \url{https://github.com/THUDM/AgentBench} & 2023-10-18 & 2025-01-30 & \buhbuh \\
\hline

LMRL Gym \cite{abdulhai_lmrl_2023} & \url{https://github.com/abdulhaim/LMRL-Gym} & 2023-11-25 & 2024-07-02 & \buhbuh \\
\hline

GameBench \cite{costarelli_gamebench_2024} & \url{https://github.com/Joshuaclymer/GameBench} & 2023-12-14 & 2024-06-27 & \buhbuh \\
\hline

AgentBoard \cite{ma_agentboard_2024} & \url{https://github.com/hkust-nlp/AgentBoard} & 2024-01-17 & 2024-04-23 & \buhbuh \\
\hline

GTBench \cite{duan_gtbench_2024} & \url{https://github.com/jinhaoduan/GTBench} & 2024-02-07 & 2024-09-07 & \buhbuh \\
\hline

ChatEval \cite{chan_chateval_2023} & \url{https://github.com/thunlp/ChatEval} & 2024-04-15 & 2024-10-19 & \buhbuh \\
\hline

TextArena \cite{guertler_textarena_2025} & \url{https://github.com/LeonGuertler/TextArena} & 2024-09-23 & 2025-06-25 & \cm \\
\hline
GuessArena \cite{yu_guessarena_2025} & \url{https://github.com/IAAR-Shanghai/GuessArena} & 2025-02-16 & 2025-05-29 & \cm \\
\hline


\end{tabular}
\caption{Game-based LLM Benchmarking Frameworks}
\label{tab:existing_frameworks}
\end{table*}